\documentclass{article}
\usepackage{ijcai20}

\usepackage{times}
\usepackage{soul}
\usepackage{url}
\usepackage[hidelinks]{hyperref}
\usepackage[utf8]{inputenc}
\usepackage[small]{caption}
\usepackage{graphicx}
\usepackage{amsmath}
\usepackage{amsthm}
\usepackage{booktabs}
\usepackage{algorithm}
\usepackage{algorithmic}
\usepackage{stfloats}
\usepackage{verbatim}
\usepackage{latexsym}
\usepackage{stfloats}
\usepackage{array}
\usepackage{bm}
\usepackage{amsfonts}
\usepackage{comment}
\usepackage{multirow}
\usepackage{bbding}
\usepackage{color}
\usepackage{CJKutf8}
\newcommand{\citep}[1]{\citeauthor{#1}~\shortcite{#1}}

\title{Modeling Voting for System Combination in Machine Translation}

\author{
Xuancheng Huang$^1$
\and
Jiacheng Zhang$^1$\and
Zhixing Tan$^1$\and
Derek F. Wong$^2$\and
Huanbo Luan$^1$\and \\
Jingfang Xu$^3$\and 
Maosong Sun$^1$\And
Yang Liu$^{1,4,5}$\footnote{Yang Liu is the corresponding author: liuyang2011@tsinghua. edu.cn.}
\affiliations
$^1$Dept. of Comp. Sci. \& Tech., BNRist Center, Institute for AI, Tsinghua University \\
$^2$NLP$^2$CT Lab / Department of Computer and Information Science, University of Macau \\
$^3$Sogou Inc.\\
$^4$Beijing Advanced Innovation Center for Language Resources \\
$^5$Beijing Academy of Artificial Intelligence
}

\begin{document}

\maketitle

\begin{abstract}
System combination is an important technique for combining the hypotheses of different machine translation systems to improve translation performance. Although early statistical approaches to system combination have been proven effective in analyzing the consensus between hypotheses, they suffer from the error propagation problem due to the use of pipelines. While this problem has been alleviated by end-to-end training of multi-source sequence-to-sequence models recently, these neural models do not explicitly analyze the relations between hypotheses and fail to capture their agreement because the attention to a word in a hypothesis is calculated independently, ignoring the fact that the word might occur in multiple hypotheses. In this work, we propose an approach to modeling voting for system combination in machine translation. The basic idea is to enable words in hypotheses from different systems to vote on words that are representative and should get involved in the generation process. This can be done by quantifying the influence of each voter and its preference for each candidate. Our approach combines the advantages of statistical and neural methods since it can not only analyze the relations between hypotheses but also allow for end-to-end training. Experiments show that our approach is capable of better taking advantage of the consensus between hypotheses and achieves significant improvements over state-of-the-art baselines on Chinese-English and English-German machine translation tasks. \footnote{We release our source code at Github: \url{https://github.com/THUNLP-MT/Voting4SC}}

\end{abstract}

\section{Introduction}

\begin{table}[!t]
    \begin{center}
        \begin{tabular}{l|l}
            \toprule
            $src$ & Ich hatte gestern einen Kuchen gegessen. \\
            \midrule
            $hyp_1$ & I ate a cake. \\
            $hyp_2$ & I eat a cakes yesterday. \\
            $hyp_3$ & I ate a fish yesterday. \\
            \midrule
            $trg$ & I ate a cake yesterday. \\
            \bottomrule
        \end{tabular}
        \caption{\label{tab:example} Example of system combination in machine translation. Given a German sentence (i.e., $src$), there are three hypotheses (i.e., $hyp_1$, $hyp_2$, and $hyp_3$) generated by three different German-English MT systems. The goal of system combination is to combine these erroneous but complementary hypotheses to obtain a better English translation (i.e., $trg$).}
    \end{center}
\end{table}

Machine translation (MT) is a challenging artificial intelligence task. Although many methods have been proposed for MT \cite{Brown:93,Koehn:03,Bahdanau:15,Vaswani2017AttentionIA}, they can only capture partial regularities of the translation process due to the complexity and diversity of natural languages. To address this problem, {\em system combination}, which aims to combine the hypotheses of multiple MT systems to obtain better translations, has been intensively studied \cite{Rosti2007ImprovedWS,he-etal-2008-indirect,Zhou2017NeuralSC} and widely used in MT evaluations \cite{barrault2019findings}. Table 1 shows an example. Given an input German sentence (i.e., $src$), there are three erroneous but complementary hypotheses (i.e., $hyp_1$, $hyp_2$, and $hyp_3$) generated by three different MT systems. The goal of system combination is to combine them to produce a better translation (i.e., $trg$).

Approaches to system combination can be roughly divided into two broad categories: \emph{statistical} and \emph{neural} approaches. Among statistical approaches combining hypotheses at different levels \cite{Bangalore2001ComputingCT,Matusov2006ComputingCT,Rosti2007CombiningOF,Rosti2007ImprovedWS,he-etal-2008-indirect,Karakos2008MachineTS,Chen2009ACS,Feng2009LatticebasedSC,Freitag2014JaneOS,Ma2015SystemCF}, word-level combination approaches \cite{Rosti2007ImprovedWS,he-etal-2008-indirect} prove to achieve the best translation performance. These approaches are capable of analyzing the relations between hypotheses and {\em  voting} on the most probable word at each position by using a pipeline: choosing a backbone, aligning the words between hypotheses, building a confusion network, and generating the translation. Despite the benefit of explicit modeling of voting, the use of pipelines often results in the error propagation problem: errors made in early steps in the pipeline will be propagated to subsequent steps.

Recently, the error propagation problem has been alleviated by end-to-end training of neural combination methods \cite{Zhou2017NeuralSC}, since system combination can be regarded as a multi-source sequence-to-sequence problem \cite{Zoph2016MultiSourceNT}.\footnote{By definition, system combination methods do not include model ensemble \cite{Xiao2013BaggingAB}, which combines the predictions of multiple homogeneous models during decoding instead of combining the hypotheses of multiple heterogeneous MT systems after decoding.} One advantage of this line of work is that the importance of a word in a hypothesis can be quantified by its attention weight, without the need for explicit hypothesis alignment. However, a key limitation is that the attention weights between one hypothesis and the  output are calculated independently, ignoring the fact that a candidate word might occur in multiple hypotheses. For example, in Table \ref{tab:example}, as ``ate'' occurs in two hypotheses and ``eat'' occurs in one hypothesis, ``ate'' should be more likely to appear in the output. However, such connections between hypotheses are not taken into consideration in existing work.

\begin{figure*}[!t]
  \centering
  \includegraphics[scale=0.27]{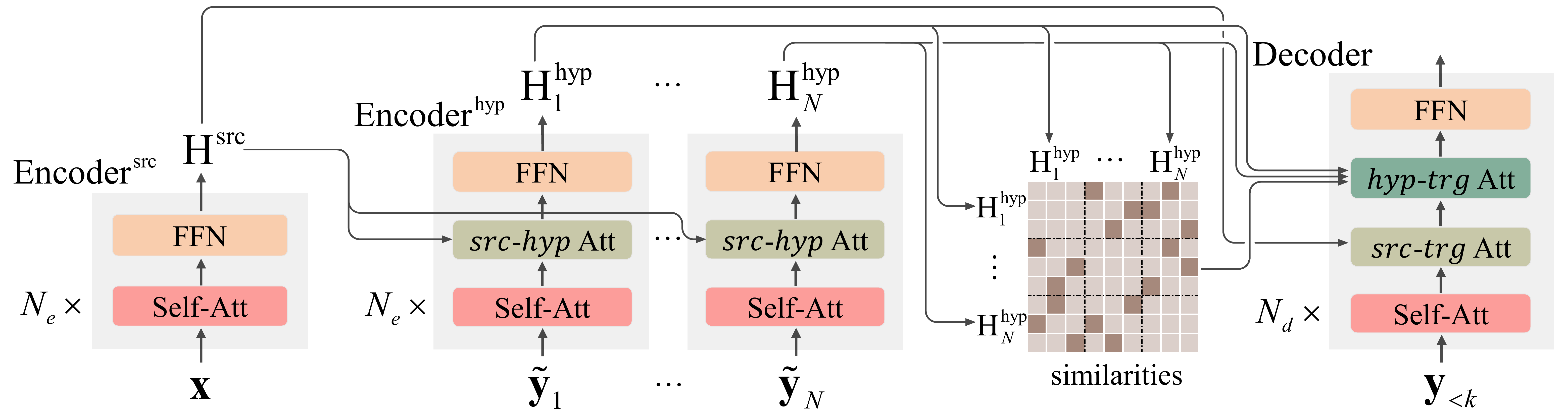}
  \caption{The architecture of our approach. Based on the multi-source sequence-to-sequence model, our approach introduces a voting mechanism to find the agreement between hypotheses. During voting, while a voter's influence depends on the source sentence $\mathbf{x}$ and the partial output $\mathbf{y}_{<k}$, its preference for a candidate depends on their word similarity calculated using the output of hypothesis encoders (i.e., $\mathbf{H}^{\mathrm{hyp}}_n$). The result of voting is used to change the attention weights between hypotheses and output (i.e., $hyp$-$trg \ \mathrm{Att}$) to encourage words receiving more votes to be more likely to be included in the output.}
  \label{fig:our_arch}
\end{figure*}

In this work, we propose an approach to model voting for system combination in machine translation. The basic idea is to find the consensus between hypotheses by enabling words in hypotheses to vote on representative words. Our approach distinguishes between two types of words during voting: {\em voter} and {\em candidate}. For example, in Table \ref{tab:example}, if ``fish'' in $hyp_3$ is chosen as a candidate, all words of $hyp_1$ and $hyp_2$ will serve as voters to decide whether ``fish'' should be included in the output. To do so, our approach quantifies the influence of each voter and its preference for each candidate using contexts and word similarities, respectively. By modifying attention weights, candidates receiving more votes are more likely to participate in the generation process. Our approach combines the merits of statistical and neural combination methods by both exploiting the relations between hypotheses and allowing for end-to-end training. Experiments show our approach achieves significant improvements over state-of-the-art statistical and neural combination methods on NIST and WMT benchmarks.

\section{Approach}

Figure \ref{fig:our_arch} shows the overall architecture of our model. The model starts with learning the representations of the source sentence and  hypotheses (Section \ref{sec:encoder}). Given the representation of hypotheses, our approach introduces voting by calculating the similarities between words in hypotheses and increasing the attention weights of similar words collectively (Section \ref{sec:voting}). Finally, the decoder takes the representations of contexts as input and encourages words on which most hypotheses agree to be more likely contribute to the generation of the output (Section \ref{sec:decoder}).

\subsection{The Encoders}
\label{sec:encoder}

As system combination takes multiple hypotheses from different MT systems as input, it is natural to cast system combination as a multi-source sequence-to-sequence problem \cite{Zoph2016MultiSourceNT} as suggested by ~\citep{Zhou2017NeuralSC}.

Formally, let $\mathbf{x}$ be a source sentence (i.e., $src$) and $\mathbf{\tilde{y}}_{1:N}=\mathbf{\tilde{y}}_{1} \dots \mathbf{\tilde{y}}_{N}$ be $N$ hypotheses generated by different MT systems, where $\mathbf{\tilde{y}}_n$ is the $n$-th hypothesis (i.e., $hyp_{n}$). We use $\tilde{y}_{n, j}$ to denote the $j$-th word of the $n$-th hypothesis. We use $\mathbf{y}=y_1 \dots y_K$ to denote the output (i.e., $trg$) with $K$ words. Hence, the system combination model is given by
\begin{alignat}{1}
P(\mathbf{y}|\mathbf{x}, \mathbf{\tilde{y}}_{1:N}; \bm{\theta}) &= \prod_{k=1}^{K}P(y_k|\mathbf{x},\mathbf{\tilde{y}}_{1:N},\mathbf{y}_{<k}; \bm{\theta}), \label{eqn:overall_prob}
\end{alignat}
where $y_k$ is the $k$-th target word, $\mathbf{y}_{<k} = y_1 \dots y_{k-1}$ is a partial output, and $\bm{\theta}$ is a set of model parameters.

To model the source sentence $\mathbf{x}$ and hypotheses $\tilde{\mathbf{y}}_{1:N}$,  our model consists of $N+1$ encoders:
\begin{alignat}{1}
\mathbf{H}^{\mathrm{src}} &= \mathrm{Encoder}^{\mathrm{src}}(\mathbf{x},\bm{\theta}), \\
\mathbf{H}^{\mathrm{hyp}}_n &= \mathrm{Encoder}^{\mathrm{hyp}}(\mathbf{\tilde{y}}_n,\bm{\theta}), 1 \le n \le N,
\end{alignat}
where $\mathrm{Encoder}^{\mathrm{src}}(\cdot)$ is the encoder for $src$, $\mathbf{H}^{\mathrm{src}}$ is the representation of $src$, $\mathrm{Encoder}^{\mathrm{hyp}}(\cdot)$ is the encoder for each hypothesis, and $\mathbf{H}^{\mathrm{hyp}}_n$ is the representation of $hyp_n$. Note that the parameters of the hypothesis encoder are shared by all hypotheses.

Although the multi-source sequence-to-sequence framework has shown the superiority of end-to-end training of neural combination methods over conventional statistical methods \cite{Zhou2017NeuralSC}, the dependencies between hypotheses that are critical for finding the consensus are ignored because symbolic hypothesis alignment is not allowed in neural networks. Therefore, it is important to re-introduce voting into modern system combination methods while keeping the benefit of end-to-end training.

\subsection{The Voting Mechanism}
\label{sec:voting}

\begin{figure}[!t]
  \centering
  \includegraphics[scale=0.415]{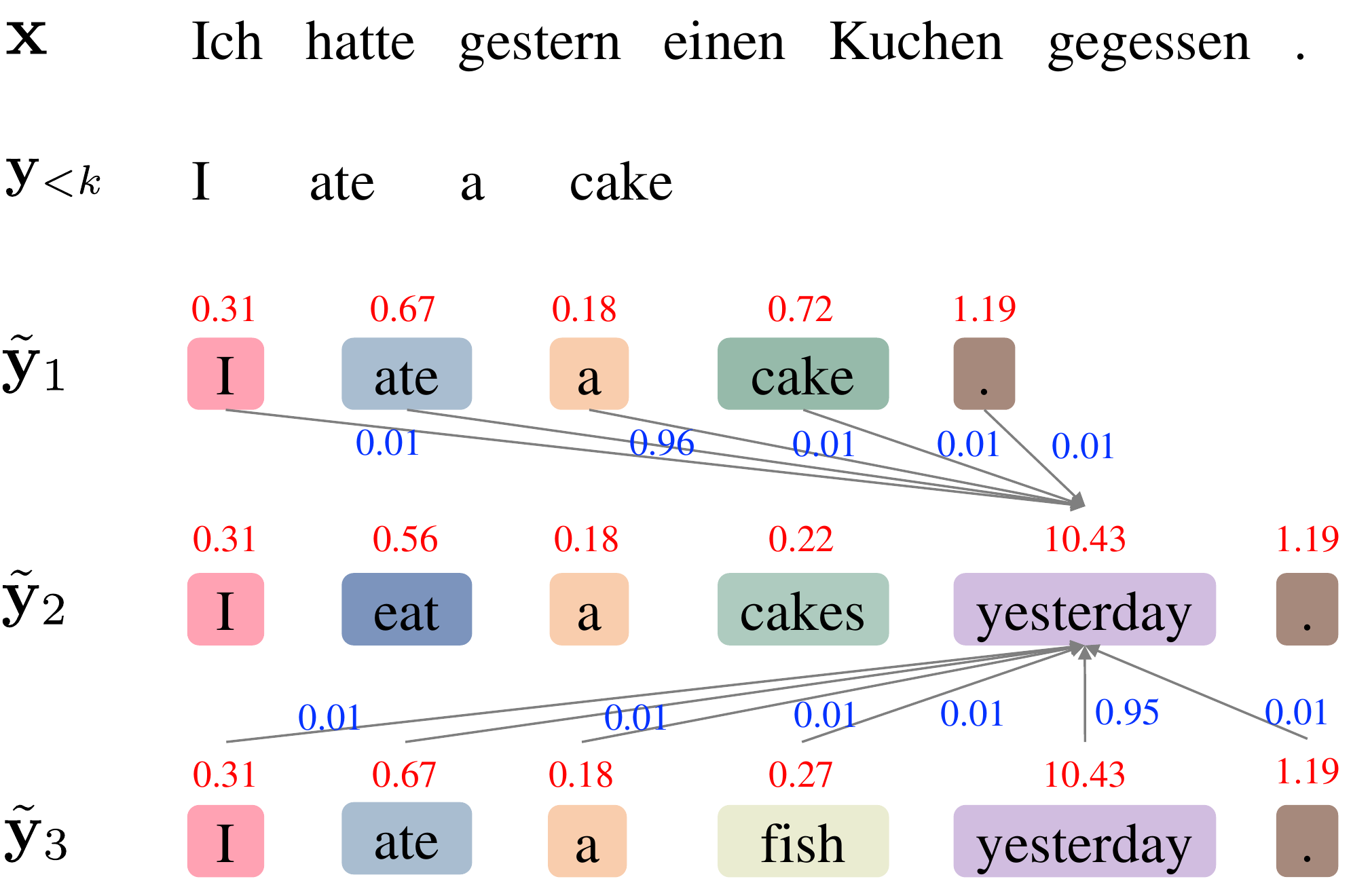}
  \caption{Voting in system combination. Similar words are highlighted in similar colors. ``yesterday'' in $\tilde{\mathbf{y}}_2$ is a candidate, which receives votes from voters in $\tilde{\mathbf{y}}_1$ and $\tilde{\mathbf{y}}_3$. The influence of each voter is a real-valued number highlighted in red and its preference for the candidate is highlighted in blue.}
  \label{fig:vote}
\end{figure}

We first use an example to illustrate the basic idea of voting. Figure \ref{fig:vote} shows a source sentence $\mathbf{x}$, a partial output $\mathbf{y}_{<k}$, and three hypotheses $\tilde{\mathbf{y}}_1$, $\tilde{\mathbf{y}}_2$, and $\tilde{\mathbf{y}}_3$. The question is which word in the hypotheses is more likely to be the fifth word of the output.

Our approach distinguishes between two kinds of words in hypotheses: {\em voter} and {\em candidate}. The voting mechanism allows each word (i.e., voter) to vote for other words (i.e., candidates) in other hypotheses. For example, the fifth word of $\tilde{\mathbf{y}}_2$ in Figure \ref{fig:vote} (i.e., ``yesterday'') is a candidate and all words in $\tilde{\mathbf{y}}_1$ and $\tilde{\mathbf{y}}_3$ are voters. The voters decide whether the candidate should be included in the output by voting.

The voting process involves two basic problems: 
\begin{enumerate}
    \item {\em Influence}: how influential is the voter? 
    \item {\em Preference}: which candidate does the voter support?
\end{enumerate}

 The influence of each voter is quantified as a real-valued number (highlighted in red in Figure \ref{fig:vote}), which is actually the energy used in calculating attention weight:
\begin{eqnarray}
e_{n, j} = f(\mathbf{x}, \mathbf{y}_{<k}, \tilde{y}_{n,j}, \bm{\theta}),
\end{eqnarray}
where $f(\cdot)$ is a function that calculates the energy, $\tilde{y}_{n, j}$ is the $j$-th word of the $n$-th hypothesis, and $e_{n, j}$ is its corresponding energy that reflects how likely it will be the next word. In Figure \ref{fig:vote}, ``yesterday'' receives the highest energy according to the source sentence and the partial output. Note that the energy of each word will change as the partial output changes. For example, the energy of ``yesterday'' will be decreased when the model is predicting the sixth word. This is much better than simply using number of occurrences that are independent of contexts as votes.

The preference of a voter for a candidate can also be measured as a real-valued number (highlighted in blue in Figure \ref{fig:vote}), which is actually the similarity between the voter and the candidate:
\begin{eqnarray}
\mathrm{sim}(\tilde{y}_{m, i}, \tilde{y}_{n, j}) = \frac{\exp(\mathbf{h}_{m, i}\mathbf{h}_{n, j}^{\top})}{\sum_{i'=1}^{|\tilde{\mathbf{y}}_m|} \exp(\mathbf{h}_{m, i'}\mathbf{h}_{n, j}^{\top}) }, \label{eqn:sim}
\end{eqnarray}
where $\tilde{y}_{m,i}$ is a voter and $\mathbf{h}_{m, i}$ is its representation retrieved from $\mathbf{H}^{\mathrm{hyp}}_m$. \footnote{We also tried other methods for calculating word similarity such as edit distance and the distance between word vectors but found that using the output of encoders works best because surrounding contexts are taken into consideration during representation learning.} Likewise, $\tilde{y}_{n,j}$ is a candidate and $\mathbf{h}_{n,j}$ is its representation. In Figure \ref{fig:vote}, ``yesterday" is mainly supported by voters ``yesterday" and ``ate". Note that the similarities between the voters and a candidate are normalized at the hypothesis level to avoid the length bias: longer hypotheses tend to send out more votes.

It is easy to collect all votes by calculating a weighted sum of energies of voters. As a result, the extended energy of each candidate can defined as
\begin{eqnarray}
\tilde{e}_{n,j} = e_{n,j} + \sum_{m=1 \land m\ne n}^{N} \sum_{i=1}^{|\tilde{\mathbf{y}}_m|} \mathrm{sim}(\tilde{y}_{m,i}, \tilde{y}_{n,j}) \times e_{m,i}.
\end{eqnarray}
Clearly, the extended energy depends on both the influence (i.e., $e_{m, i}$) and the preference (i.e., $\mathrm{sim}(\tilde{y}_{m,i}, \tilde{y}_{n,j})$).

Finally, the result of voting is used to change the attention weights between the hypotheses and output:
\begin{alignat}{1}
\alpha_{n,j} &= \frac{\mathrm{exp}(\tilde{e}_{n,j})}
{\sum_{n'=1}^{N}\sum_{j'=1}^{|\tilde{\mathbf{y}}_{n'}|}{\mathrm{exp}(\tilde{e}_{n',j'})}}. \label{eqn:alpha}
\end{alignat}
In this way, candidates receiving strong support such as ``yesterday" in Figure \ref{fig:vote} will be more likely to appear in the output. 

\subsection{The Decoder}
\label{sec:decoder}

The decoder takes the representations of contexts as input and generates the output:
\begin{alignat}{1}
\mathbf{h}^{\mathrm{trg}}_k &= \mathrm{Decoder}(\mathbf{y}_{<k}, \mathbf{H}^{\mathrm{src}}, \mathbf{H}^{\mathrm{hyp}}_{1:N}, \bm{\theta}), \label{eqn:h_pe}\\
P(y_k&|\mathbf{x},\mathbf{\tilde{y}}_{1:N},\mathbf{y}_{<k}; \bm{\theta}) \propto g(\mathbf{h}^{\mathrm{trg}}_k),
\end{alignat}
where $\mathrm{Decoder}(\cdot)$ is the decoder, $\mathbf{h}^{\mathrm{trg}}_k$ is the representation of the $k$-th target word $y_k$ in $trg$, and $g(\cdot)$ is a function that calculates the generation probabilities. The difference between our $\mathrm{Decoder}(\cdot)$ and the decoder described in Transformer~\cite{Vaswani2017AttentionIA} is that our $\mathrm{Decoder}(\cdot)$ contains additional $hyp$-$trg$ attention layers, which replace the attention weight with $\alpha_{n,j}$ described in Eq.(\ref{eqn:alpha}) (See Figure \ref{fig:our_arch}) .

When designing the decoder, we take advantage of two characteristics of system combination. First, words in hypotheses account for a large portion of those in the ground-truth output. Second, the hypotheses and output belong to the same language, making it convenient to detect and reduce word omission and repetition.

\subsubsection*{Using both Restricted and Full Vocabularies}
\label{sec:vocab}

The first characteristic of system combination is that most words in hypotheses will appear in the ground-truth output. This is why conventional statistical combination methods \cite{Rosti2007ImprovedWS} only use a restricted vocabulary which contains words in hypotheses to constrain the search space. However, as there are still words in the ground-truth output falling outside the restricted vocabulary, modern neural combination methods often use the full vocabulary to ensure the coverage. Nevertheless, the observation that words in the hypotheses are more likely to be included in the output is not fully exploited.

As a result, we propose to use both restricted and full vocabularies to both take advantage of the observed regularity and ensure coverage. This can be done by calculating the output probabilities using restricted and full vocabularies separately and interpolating them using a gate:
\begin{eqnarray}
&& P(y_k|\mathbf{x},\mathbf{\tilde{y}}_{1:N},\mathbf{y}_{<k}; \bm{\theta}) \nonumber \\
&=& \lambda_k \times P_r(y_k|\mathbf{x},\mathbf{\tilde{y}}_{1:N},\mathbf{y}_{<k}; \bm{\theta}) + \nonumber \\
&& (1- \lambda_k) \times P_f(y_k|\mathbf{x},\mathbf{\tilde{y}}_{1:N},\mathbf{y}_{<k}; \bm{\theta}),
\end{eqnarray}
where $P_r(\cdot)$ is the probability calculated using the restricted vocabulary, $P_f(\cdot)$ is the probability calculated using the full vocabulary, and $\lambda_k$ is a gate for predicting $y_k$. Note that $\lambda_k$ obtained by $\mathbf{h}^{\mathrm{trg}}_k$ after a simple linear transformation and Sigmoid activation function, which depends on the context and is not fixed during decoding.

\subsubsection*{Improving Coverage using Dynamic Weighting}
\label{sec:decoding}
Due to the lack of coverage vector that indicates whether a source word is translated or not \cite{Koehn:03}, word omission and repetition are severe problems in neural MT systems that hinder translation performance \cite{Tu2016ModelingCF}. Fortunately, in system combination the hypotheses and output are in the same language, making it possible to detect and reduce word omission and repetition. For example, in Figure \ref{fig:vote}, as ``cake'' has already been included in the output, it should not appear in the output again. We propose to assign each word in hypotheses a weight, which changes dynamically during decoding: once a word is included in the output, its weight is decreased accordingly.

More precisely, we count the frequency of the words in hypotheses on average, which is represented by $\mathbf{c}^h \in \mathbb{R}^{|\mathcal{V}_f|}$. Meanwhile, we count the frequency of the words in $\mathbf{y}_{<k}$, which is denoted as $\mathbf{c}^y_k \in \mathbb{R}^{|\mathcal{V}_f|}$. Then, the probability distribution can be weighted by:
\begin{alignat}{1}
\mathbf{w}_k &= \log_2\big(\max(\mathbf{c}^h-\mathbf{c}_k^y, 0)+2\big), \\
\tilde{\mathbf{p}}_k &\propto \mathbf{w}_k \odot \mathbf{p}_k,
\end{alignat}
where $\mathbf{w}_k \in \mathbb{R}^{|\mathcal{V}_f|}$ is a weight vector, $\tilde{\mathbf{p}}_k \in \mathbb{R}^{|\mathcal{V}_f|}$ is a weighted probability distribution for $y_k$, $\mathcal{V}_f$ is the full vocabulary, and $\odot$ is the point-wise multiplication. We keep the weight no smaller than 1.0 for each word and use a logarithmic function to deal with unexpected high frequency. At last, $\tilde{\mathbf{p}}_k$ is utilized by the beam-search algorithm to select top partial translations.

\section{Experiments}
\subsection{Setup}
\subsubsection*{Datasets}
We evaluated our approach on Chinese-English (Zh-En) and English-German (En-De) translation tasks. For the Chinese-English task, the training set contains about 1.25M sentence pairs from LDC with 27.9M Chinese words and 34.5M English words. \footnote{The training set includes LDC2002E18, LDC2003E07, LDC-2003E14, part of LDC2004T07, LDC2004T08 and LDC2005T06.} We used the NIST 2006 dataset as the development set. The NIST 2002, 2003, 2004, 2005, and 2008 datasets were used as test sets. For the English-German task, the training set is the WMT 2014 training data with 4.5M sentence pairs, the validation set is newstest2013, and the test set is newstest2014. We used byte pair encoding (BPE) \cite{Sennrich2016NeuralMT} with 32K merges to segment words into sub-word units.

\subsubsection*{Evaluation Metrics}
We report case-insensitive tokenized BLEU scores for Chinese-English and case-sensitive tokenized BLEU scores for English-German. For English-German, we apply compound splitting\footnote{\url{https://github.com/pytorch/fairseq/blob/master/scripts/compound\_split\_bleu.sh}} similar to that of \citep{Vaswani2017AttentionIA}. We used the paired bootstrap resampling \cite{Koehn2004StatisticalST} for statistical significance tests.

\subsubsection*{Single MT Systems}
For the system combination task, we constructed all training datasets with outputs of three translation systems. We used the same systems for training, validation, and testing for both language pairs. Except for pretrained MT model, all single MT systems used the same training dataset as combination systems for training. 

For the Chinese-English task, we used Transformer-base~\cite{Vaswani2017AttentionIA} with left-to-right decoding  (\textsc{Trans-L2R}), Transformer-base with right-to-left decoding (\textsc{Trans-R2L}) and non-autoregressive translation model (\textsc{Mask-NAT})~\cite{Ghazvininejad2019MaskPredictPD} as the three ``black-box" translation systems. We used the training data simulation strategy~\cite{Zhou2017NeuralSC} to alleviate the training-test bias. 

For the English-German task, in order to demonstrate the effectiveness of our approach on top of the state-of-the-art results, we used FAIR's pretrained Transformer-big (\textsc{Transformer$_\mathrm{big}$-fb}) \cite{Ott2018ScalingNM}, DynamicConv (\textsc{DynamicConv}) \cite{Wu2019PayLA}, and vanilla Transformer-big (\textsc{Transformer}$_\mathrm{big}$) \cite{Vaswani2017AttentionIA} as individual MT systems.

\begin{table*}[!t]
    \begin{center}
        \begin{tabular}{l|lllll|l}
            \toprule
            Method & NIST02 & NIST03 & NIST04 & NIST05 & NIST08 & \ \,All\\
            \midrule
            \textsc{Trans-R2L} & 45.11 & 44.67 & 46.66 & 46.08 & 36.90 & 44.26 \\
            \textsc{Mask-NAT}~{\tiny \cite{Ghazvininejad2019MaskPredictPD}}& 46.69 & 45.93 & 47.27 & 45.72 & 36.14 & 44.76 \\
            \textsc{Trans-L2R}~{\tiny \cite{Vaswani2017AttentionIA}} & 47.25 & 47.30 & 47.97 & 47.64 & 37.49 & 45.92 \\
            \midrule
            \textsc{Jane}~{\tiny \cite{Freitag2014JaneOS}} & 47.75 & 47.88 & 48.90 & 48.83 & 38.66 & 46.78 \\
            \textsc{Hier}~{\tiny \cite{Zhou2017NeuralSC}} & 48.71 & 48.31 & 48.96 & 48.74 & 38.42 & 46.85 \\
            \textsc{Ours} & \textbf{49.30}$^{\dag\dag\ddag\ddag**}$ & \textbf{49.24}$^{\dag\dag\ddag\ddag**}$ & \textbf{49.65}$^{\dag\dag\ddag\ddag**}$ & \textbf{49.28}$^{\dag\dag\ddag**}$ & \textbf{39.41}$^{\dag\dag\ddag\ddag**}$ & \textbf{47.69}$^{\dag\dag\ddag\ddag**}$ \\
            \bottomrule
        \end{tabular}
        \caption{\label{tab:ch-en} Results on the Chinese-English task. The evaluation metric is case-insensitive tokenized BLEU. ``All" is the concatenation of all test sets. The translations of the top three single MT systems are the inputs of the bottom three system combination methods. 
        ``\dag\dag": significantly better than ``\textsc{Transformer-L2R}" ($p < 0.01$). ``\ddag" and ``\ddag\ddag": significantly better than ``\textsc{Jane}" ($p < 0.05$ and $p < 0.01$). ``**": significantly better than ``\textsc{Hier}" ($p < 0.01$).}
    \end{center}
\end{table*}

\subsubsection*{Baselines}
We compared our approach with the following two state-of-the-art statistical and neural combination methods:
\begin{enumerate}
    \item \textsc{Jane}~\cite{Freitag2014JaneOS}: a statistical system combination method included in RWTH’s open-source statistical machine translation toolkit. It was designed based on confusion networks \cite{Rosti2007ImprovedWS,he-etal-2008-indirect}. We used \textsc{Jane} with its default setting. 
    \item \textsc{Hier}~\cite{Zhou2017NeuralSC}: a neural system combination method leveraging multi-source NMT framework~\cite{Zoph2016MultiSourceNT}. It was originally designed for the RNNsearch model \cite{Bahdanau:15}. To make a fair comparison, we adapted it to the Transformer model in our experiments since our approach is built on top of Transformer.
\end{enumerate}

\subsubsection*{Hyper-parameter Setting}
We used the same hyper-parameter setting for both baselines and our approach. The number of layers was set to 6 for both encoders and decoder. The hidden size was set to 512 and the filter size was set to 2,048. The number of individual attention heads was set to 8 for multi-head attention. We tied all three $src, hyp, trg$ embeddings for the English-German task. The embeddings and softmax weights were tied for both language pairs. In training, we used Adam \cite{Kingma2015AdamAM} for optimization. Each mini-batch contains 19K tokens for the Chinese-English task and 25K tokens for the English-German task. We used the learning rate decay policy described by \cite{Vaswani2017AttentionIA}. In decoding, the beam
size was set to 4 for both language pairs and the length penalty was set to 1.0 and 0.6 for Chinese-English and English-German, respectively. The other hyper-parameter settings were the same as the Transformer-base model \cite{Vaswani2017AttentionIA}. We used the development set to select the best model. We implemented our approach on top of the opensource toolkit THUMT\footnote{\url{https://github.com/THUNLP-MT/THUMT}}.

\begin{table}[!t]
    \begin{center}
        \begin{tabular}{l|l}
            \toprule
            Method &  newstest2014 \\
            \midrule
            \textsc{Transformer}$_\mathrm{big}$~{\tiny \cite{Vaswani2017AttentionIA}}& 28.72 \\
            \textsc{DynamicConv}~{\tiny \cite{Wu2019PayLA}} & 29.74 \\
            \textsc{Transformer$_\mathrm{big}$-fb}~{\tiny ~\cite{Ott2018ScalingNM}} & 29.76 \\
            \midrule
            \textsc{Jane}~{\tiny \cite{Freitag2014JaneOS}} & 29.62 \\
            \textsc{Hier}~{\tiny \cite{Zhou2017NeuralSC}} & 29.95 \\
            \textsc{Ours} & \textbf{30.52}$^{\dag\dag\ddag\ddag**}$ \\
            \bottomrule
        \end{tabular}
        \caption{\label{tab:wmt} Results on the English-German task. The evaluation metric is case-sensitive tokenized BLEU.  The translations of the top three single MT systems are the inputs of the bottom three system combination methods. ``\dag\dag": significantly better than ``\textsc{Transformer$_\mathrm{big}$-fb}" ($p < 0.01$).  ``\ddag\ddag": significantly better than ``\textsc{Jane}" ($p < 0.01$). ``**": significantly better than ``\textsc{Hier}" ($p < 0.01$).}
    \end{center}
\end{table}

\begin{table}[!t]
    \begin{center}
        \begin{tabular}{l|c}
            \toprule
            Model &  NIST06 \\
            \midrule
            \textsc{Ours} & \textbf{49.00} \\
            \midrule
            $-$~Dynamic Weighting & 48.72 \\
            $-$~Restricted Vocabulary & 48.65 \\
            $-$~Voting Mechanism  & 48.55 \\
            \bottomrule
        \end{tabular}
        \caption{\label{tab:ablation} Ablation study. ``Dynamic Weighting" denotes improving coverage using dynamic weighting (see Section \ref{sec:decoding}), ``Restricted Vocabulary" denotes using both restricted and full vocabularies (see Section \ref{sec:vocab}), and ``Voting Mechanism” denotes the voting mechanism (see Section \ref{sec:voting}).}
    \end{center}
\end{table}

\subsection{Main Results}
Table \ref{tab:ch-en} shows the results on the Chinese-English task. We find that our method outperforms the best single system \textsc{Trans-L2R}, the statistical combination method \textsc{Jane}, and the neural combination method \textsc{Hier}. All the differences are statistically significant ($p<0.01$). The superiority over \textsc{Jane} and \textsc{Hier} suggests that combining the merits of analyzing the dependencies between hypotheses and end-to-end training of neural networks helps to generate better translations.

Table \ref{tab:wmt} shows the results on the English-German task. Our approach also achieves significant improvements over the state-of-the-art results ($p < 0.01$). The gaps are relatively smaller than those on Chinese-English because the English-German task uses single reference while Chinese-English uses four references. Considering that \textsc{Transformer$_\mathrm{big}$-fb} is nowadays acknowledged strongest single system result, \textsc{Jane} cannot improve the translation quality and \textsc{Hier} improves a little while our approach achieves significant improvements, indicating that voting mechanism and end-to-end training of neural networks is important especially when the translations of single systems already have high quality.

\begin{table*}[!t]
    \begin{center}
        \begin{tabular}{l|ccccc|c|c||cl|c}
            \toprule
            $N$ & \multicolumn{5}{c|}{Single MT Systems} & Training & Test &  \textsc{Hier} & \textsc{Ours} & $\Delta$ \\
            \midrule
            \;2     & 47.44 & 46.18 &   -   &   -   &   -   & No & Yes &  47.35 & \textbf{47.76}$^{\dag\ddag}$ & +0.41 \\
            \;3 & 47.44 & 46.18 & 45.18 &   -   &   -   & Yes & Yes &  48.26 & \textbf{49.00}$^{\dag\dag\ddag\ddag}$ & +0.74 \\
            \;4     & 47.44 & 46.18 & 45.18 & 47.09 &   -   & No & Yes &  48.59 & \textbf{49.56}$^{\dag\dag\ddag\ddag}$ & +0.97 \\
            \;5     & 47.44 & 46.18 & 45.18 & 47.09 & 45.97 & No & Yes & 48.79 & \textbf{49.80}$^{\dag\dag\ddag\ddag}$ & +1.01 \\
            \bottomrule
        \end{tabular}
        \caption{\label{tab:analysis} Generalization ability evaluation. We report the results of single MT systems and  system combination methods (i.e. \textsc{Hier} and \textsc{Ours}) on the Chinese-English task. $N$ denotes the number of single MT systems. Note that both system combination methods were trained on the outputs of three single MT systems (i.e., $N=3$) and tested on various number of single systems (i.e., $N=2,3,4,5$). ``\dag" and ``\dag\dag": significantly better than the best system among inputs ($p < 0.05$ and $p < 0.01$). ``\ddag" and ``\ddag\ddag": significantly better than ``\textsc{Hier}" ($p < 0.05$ and $p < 0.01$).}
    \end{center}
\end{table*}

\subsection{Ablation Study}
\label{sec:ablation}
Table \ref{tab:ablation} shows the results of ablation study. We find that the voting mechanism seems to play a critical role since removing it deteriorates the translation performance, which can be attributed to finding the consensus between hypotheses. Both combining restricted and full vocabularies and dynamic weighting are shown to be beneficial for improving system combination but seem to have relatively smaller contributions than the voting mechanism.

\subsection{Generalization Ability Evaluation}
\label{sec:analysis}
Table \ref{tab:analysis} shows the results of generalization ability evaluation. We are interested in evaluating whether a system combination method performs well when the number of single MT systems during testing is different from that during training.  

In this experiment, we used the outputs of three single systems for training combination methods and tested them on two, three, four, and five single systems. \footnote{As \textsc{Jane} requires that the number of single systems during testing must be identical to that during training, it was not included in this experiment.} We find that our approach consistently improves over single systems while \textsc{Hier} underperforms the best single system when $N=2$. In addition, the gap between our approach and \textsc{Hier} grows with the increase of $N$, indicating that the more hypotheses, the more effective the voting mechanism. 

Specifically, we found that the number of single MT systems are more important than the averaged performance of single MT systems (See $N=4$ and $N=5$ in Table \ref{tab:analysis}), indicating that the consensus of different hypotheses are more important than the quality of individual hypothesis. We shared the parameters of each hypothesis encoder $\mathrm{Encoder}^{\mathrm{hyp}}(\cdot)$ so that the system combination model has the ability to generalize to different number of single systems.

\begin{table}[!t]
    \begin{center}
        \begin{tabular}{r|cc}
            \toprule
            Rank &  \multicolumn{2}{c}{Word Pair} \\
            \midrule
            37 & kind & type \\
            84 & example & instance \\
            96 & made & makes \\
            97 & besides & except \\
            151 & greatest & broadest \\
            156 & make & makes \\
            165 & difficulties & difficulty \\
            211 & strategic & strategy \\
            224 & moreover & furthermore \\
            228 & 7 & 7@@ \\
            \bottomrule
        \end{tabular}
        \caption{\label{tab:word-pair} The top-10 similar non-identical word pairs. ``Rank" refers to the rank among all word pairs, which is calculated according to their similarities in our approach (See Eq.(\ref{eqn:sim})).}
    \end{center}
\end{table}

\subsection{Effectiveness of Similarity Calculation}
\label{sec:word-pair}
Table \ref{tab:word-pair} shows the top-10 similar non-identical word pairs according to their similarities (See Eq.(\ref{eqn:sim})). The similarity are averaged by every occurrences of the word pair on the Chinese-English development set. The higher the ranking, the higher the similarity.  We only display word pairs that two words are non-identical because the similarity between two identical words are naturally high. However, we find that some non-identical word pairs have signally high similarities (e.g., the similarity between ``kind'' and ``type'', which is the highest ranked non-identical word pair, is ranked 37, higher than most of identical word pairs). It can be attributed to that different words might have similar meanings, indicating that our approach is able to catch the similar words in spite of their different morphologies.

\begin{figure*}[!t]
  \centering
  \includegraphics[scale=0.7]{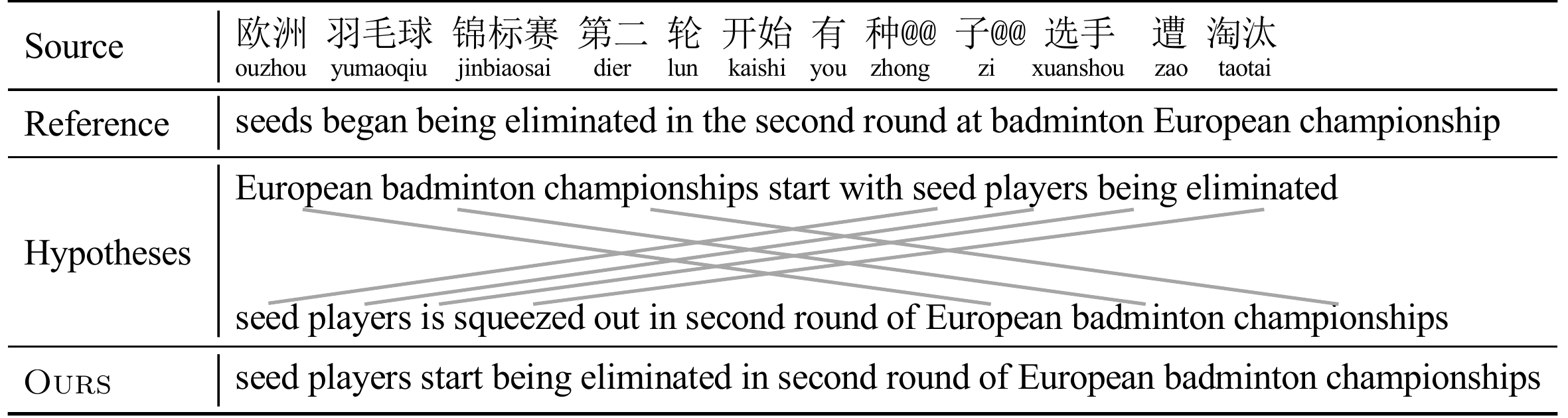}
  \caption{Case study. Word pairs with high similarities are aligned with lines. We find that our approach is able to identify similar words between hypotheses even if they differ in word orders significantly.}
  \label{fig:case}
\end{figure*}

Figure \ref{fig:case} shows an example on using our approach to combine two single MT systems on the Chinese-English task. Word pairs with similarities higher than 0.12 are aligned with lines. We find that our approach is capable of identifying similar words such as ``eliminated'' and ``squeezed'' even if the two hypotheses have significantly different syntactic structures.

\begin{figure}[!t]
  \centering
  \includegraphics[scale=0.585]{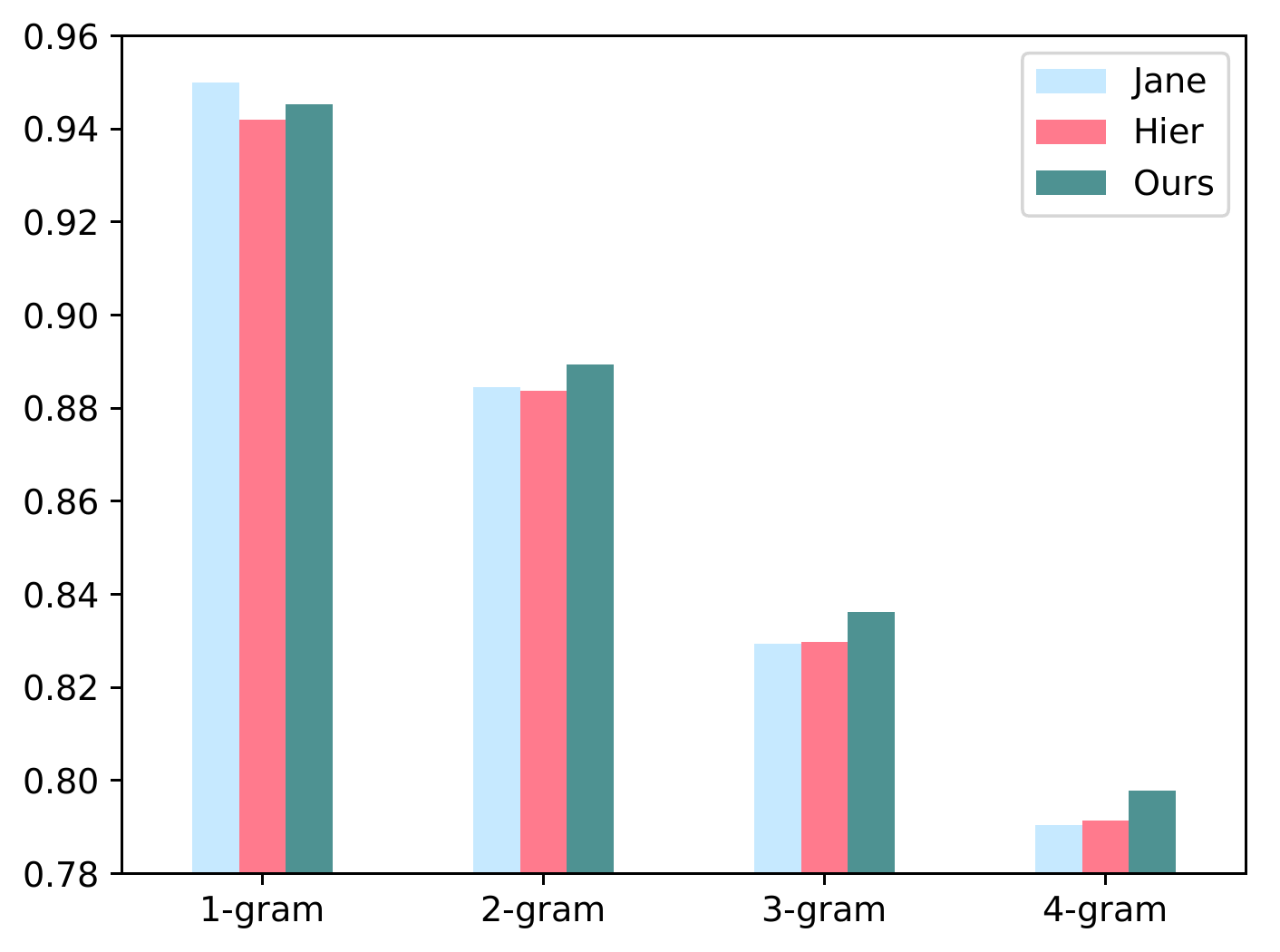}
  \caption{Averaged matching rate between outputs and hypotheses. We calculates matching rates of each hypothesis and output and averages them in $n$-grams ($n$=1, 2, 3 and 4), respectively, which is calculated on the Chinese-English development set.}
  \label{fig:ave_bleu}
\end{figure}

\subsection{Matching Rate Evaluation}
We attempt to evaluate whether our approach really leverage the consensus between hypotheses. Figure \ref{fig:ave_bleu} shows the averaged matching rate between outputs and hypotheses in $n$-grams ($n$=1, 2, 3 and 4), respectively. The matching rate for individual $n$-gram between single hypothesis and output is the same as the matching rate in BLEU metric. The averaged one are averaged among different hypotheses, which reflects weather the output adopted the consensuses of hypotheses. As shown in this figure, \textsc{Jane} achieves highest averaged matching rate in 1-gram but as the interval becomes lager, its averaged matching rate decreases. Our method achieves highest averaged matching rate in $n$-grams ($n>1$), which can be attributed to the voting mechanism and end-to-end training of neural architecture. 

\section{Related Work}

System combination has been an active field in the past two decades and there are numerous valuable literatures addressing it from different aspects   \cite{Bangalore2001ComputingCT,Matusov2006ComputingCT,Rosti2007CombiningOF,Rosti2007ImprovedWS,Huang2007HierarchicalSC,he-etal-2008-indirect,Li2008WordRA,Karakos2008MachineTS,Feng2009LatticebasedSC,Chen2009ACS,Du2010UsingTT,Heafield2010CombiningMT,Ma2012PhraselevelSC,Freitag2014JaneOS,Ma2015SystemCF,Freitag2015LocalSV,Zhu2016SentenceLevelPF,Zhou2017NeuralSC,barrault2019findings}. 

Our idea of introducing voting into neural combination methods is inspired by classical confusion networks \cite{Rosti2007ImprovedWS,he-etal-2008-indirect}, which leverage word alignment between hypotheses to find groups of competing candidate words. The voting mechanism proposed in this work is significantly different from confusion networks in two aspects. First, our approach does not rely on a pipeline involving symbolic structures and thus facilitates end-to-end training of neural networks. This is important for alleviating the error propagation problem. Second, our approach leverages the contexts to compute votes dynamically. In other words, the influence and preference of a voter will change as the partial output changes. 

The neural combination method proposed by \citep{Zhou2017NeuralSC} first shows the effectiveness of end-to-end training of multi-source sequence-to-sequence models in system combination. Along this direction, our work is also based on the same framework but focuses more on introducing voting into system combination. Our work shows that it is important to analyze the relations between hypotheses to find their consensus. As the attention weight between the hypothesis and output is calculated separately, we propose to increase the attention weights of a group of identical or similar words receiving high votes collectively. 

Our work is also related to model ensemble \cite{Xiao2013BaggingAB} widely used in the deep learning community. The major difference is that model ensemble is a ``white-box'' method that aims to integrate predictions of multiple homogeneous models during inference while system combination is a ``black-box'' method that tries to combine hypotheses of multiple heterogeneous systems after inference. 

\section{Conclusion}
We have presented a voting mechanism for system combination in machine translation. Our approach combines the advantages of statistical and neural methods by taking the relations between hypotheses into account and training models in an end-to-end manner. The voting mechanism allows words in hypotheses to vote on words that should be included in the output. Experiments show our approach achieves significant improvements over state-of-the-art baselines on Chinese-English and English-German translation tasks. 

\section*{Acknowledgements}
This work was supported by  the  National  Key  R\&D  Program  of  China (No.2017YFB0202204),  National  Natural  Science Foundation of China (No.61925601, No.61761166008, No.61772302), Beijing Advanced Innovation Center for Language Resources (No.TYR17002), and the NExT++ project supported by the National Research Foundation, Prime Ministers Office, Singapore under its IRC@Singapore Funding Initiative. Derek F. Wong is supported by the National Natural Science Foundation of China (Grant No. 61672555), the Science and Technology Development Fund, Macau SAR (Grant Nos. 045/2017/AFJ, 0101/2019/A2).

\newpage


\bibliographystyle{named}
\bibliography{ijcai20}

\end{document}